\PassOptionsToPackage{numbers,compress}{natbib}
\documentclass{article}

\usepackage[preprint]{neurips_2020}

\usepackage[utf8]{inputenc} 
\usepackage[T1]{fontenc}    
\usepackage[colorlinks]{hyperref}       
\usepackage{url}            
\usepackage{booktabs}       
\usepackage{amsfonts}       
\usepackage{nicefrac}       
\usepackage{microtype}      
\usepackage{wrapfig}        

\usepackage{mathtools,amssymb,amsthm,algorithm,algpseudocode}
\usepackage{chngcntr}

\DeclareMathOperator{\sign}{sign}

\newcommand{\pkg}[1]{\texttt{#1}}

\newtheorem{theorem}{Theorem}
\newtheorem{proposition}{Proposition}

\newtheorem{example}{Example}

\newtheorem{remark}{Remark}
\theoremstyle{remark}
\newtheorem{case}{Case}
\counterwithin*{case}{theorem}
\counterwithin*{case}{corollary}
\counterwithin*{case}{proposition}
\counterwithin*{case}{subsection}
\algdef{SE}[DOWHILE]{Do}{doWhile}{\algorithmicdo}[1]{\algorithmicwhile\ #1}

\usepackage{multirow}
\usepackage{makecell}

\title{The Strong Screening Rule for SLOPE}

\author{%
  Johan~Larsson\\
  Dept. of Statistics, Lund University\\
  \texttt{johan.larsson@stat.lu.se}\\
  \And
  Małgorzata~Bogdan\\
  Dept. of Mathematics, University of Wroclaw\\
  Dept. of Statistics, Lund University\\
  \texttt{malgorzata.bogdan@uwr.edu.pl}\\
  \And
  Jonas~Wallin\\
  Dept. of Statistics, Lund University\\
  \texttt{jonas.wallin@stat.lu.se}\\
}

\begin{document}

\maketitle

\begin{abstract}
  Extracting relevant features from data sets where the number of observations (\(n\)) is much smaller then the number of predictors (\(p\)) is a major challenge in modern statistics. Sorted L-One Penalized Estimation (SLOPE)---a generalization of the lasso---is a promising method within this setting. Current numerical procedures for SLOPE, however, lack the efficiency that respective tools for the lasso enjoy, particularly in the context of estimating a complete regularization path. A key component in the efficiency of the lasso is predictor screening rules: rules that allow  predictors to be discarded before estimating the model. This is the first paper to establish such a rule for SLOPE. We develop a screening rule for SLOPE by examining its subdifferential and show that this rule is a generalization of the strong rule for the lasso. Our rule is heuristic, which means that it may discard predictors erroneously. In our paper, however, we show that such situations are rare and easily safeguarded against by a simple check of the optimality conditions. Our numerical experiments show that the rule performs well in practice, leading to improvements by orders of magnitude for data in the \(p \gg n\) domain, as well as incurring no additional computational overhead when \(n > p\).
\end{abstract}

\hypertarget{introduction}{%
  \section{Introduction}\label{introduction}}

Extracting relevant features from data sets where the number of observations (\(n\)) is much smaller then the number of predictors (\(p\)) is one of the major challenges in modern statistics. The dominating method for this problem, in a regression setting, is the lasso~\citep{tibshirani1996}. Recently, however, an alternative known as Sorted L-One Penalized Estimation (SLOPE) has been proposed~\citep{bogdan2013,bogdan2015,zeng2014a}, which is a generalization of the Octagonal Shrinkage and Supervised Clustering Algorithm for Regression (OSCAR)~\citep{bondell2008}.


SLOPE is a regularization method that uses the sorted \(\ell_1\) norm instead of the regular \(\ell_1\) norm, which is used in the lasso. SLOPE features several interesting properties, such as control of the false discovery rate~\citep{bogdan2013,brzyski2018}, asymptotic minimaxity~\citep{su2016}, and clustering of regression coefficients in the presence of strong dependence between predictors~\citep{zeng2014}.


In more detail, SLOPE solves the convex optimization problem
\begin{equation}
  \text{minimize}_{\beta \in \mathbb{R}^p}
  \left\{ f(\beta) + J(\beta;\lambda) \right\},
  \label{eq:primal}
\end{equation}
where \(f(\beta)\) is smooth and convex and \(J(\beta;\lambda) = \sum_{j=1}^p \lambda_j \lvert \beta \rvert_{(j)}\) is the convex but non-smooth sorted \(\ell_1\) norm~\citep{bogdan2013,zeng2014}, where \(|\beta|_{(1)} \geq |\beta|_{(2)} \geq \cdots \geq |\beta|_{(p)}\) and \(\lambda_1 \geq \dots \geq \lambda_p \geq 0\).

It is easy to see that the lasso is a specific instance of SLOPE, obtained by setting all elements of \(\lambda\) to the same value. But in contrast to SLOPE based on a decreasing sequence \(\lambda\), the lasso suffers from unpredictable behavior in the presence of highly correlated predictors, occasionally resulting in solutions wherein only a subset among a group of correlated predictors is selected.
SLOPE, in contrast, turns out to be robust to correlated designs, which it accomplishes via clustering: setting coefficients of predictors to the same value~\citep{zeng2014}. \citet{kremer2019} showed that this clustering is related to similarity of the influence of respective variables on the likelihood function, which can occur due to strong correlation~\citep{zeng2014,figueiredo2016} but also due to similarity of true regression coefficients~\citep{schneider2020}. This property in some cases allows SLOPE to select all $p$ predictors if they are grouped into no more than $n$ clusters~\citep{kremer2019,schneider2020}, while the lasso can at most select \(n\) predictors~\citep{efron2004}.

The choice of \(\lambda\) sequence in \eqref{eq:primal} typically needs to be based on cross-validation or similar schemes. Most algorithms for fitting sparse regression, such as as the one implemented for lasso in the \pkg{glmnet} package for R~\citep{friedman2007}, accomplish this by constructing a path of decreasing \(\lambda\). For SLOPE, we begin the path with \(\lambda^{(1)}\) and finish at \(\lambda^{(l)}\) with \(\lambda^{(m)}_j \geq \lambda^{(m+1)}_j\) for \(j = 1,2,\dots,p\) and \(m=1,2,\dots,l-1\). (See \autoref{setup} for details regarding the construction of the path.) For any point along this path, we let \(\hat\beta(\lambda^{(m)})\) be the respective SLOPE estimate, such that
\[
  \hat\beta(\lambda^{(m)}) = \mathop{\mathrm{arg\,min}}_{\beta\in\mathbb{R}^p}\left\{f(\beta) + J(\beta;\lambda^{(m)})\right\}.
\]
Fitting the path repeatedly by cross-validation introduces a heavy computational burden. For the lasso, an important remedy for this issue arose with the advent of screening rules, which provide criteria for discarding predictors before fitting the model.

Screening rules can be broken down into two categories: \emph{safe} and \emph{heuristic} (unsafe) screening rules. The former of these guarantee that any predictors screened as inactive (determined to be zero by the rule) are in fact zero at the solution. Heuristic rules, on the other hand, may lead to \emph{violations:} incorrectly discarding predictors, which means that heuristic rules must be supplemented with a check of the Karush--Kuhn--Tucker (KKT) conditions. For any predictors failing the test, the model must be refit with these predictors added back in order to ensure optimality.

Safe screening rules include the safe feature elimination rule (SAFE~\citep{elghaoui2010}), the dome test~\citep{xiang2012}, Enhanced Dual-Polytope Projection (EDPP~\citep{wang2015}), and the Gap Safe rule~\citep{ndiaye2017,fercoq2015}. Heuristic rules include Sure Independence Screening (SIS~\citep{fan2008}), Blitz~\citep{johnson2015}, and the strong rule~\citep{tibshirani2012}. There have also been attempts to design \emph{dynamic} approaches to screening~\citep{bonnefoy2014} as well as hybrid rules, utilizing both heuristic and safe rules in tandem~\citep{zeng2021}.


The implications of screening rules have been remarkable, allowing lasso models in the \(p \gg n\) domain to be solved in a fraction of the time required otherwise and with a much reduced memory footprint~\citep{tibshirani2012}. Implementing screening rules for SLOPE has, however, proven to be considerably more challenging. After the first version of this paper appeared on \emph{arXiv}~\citep{larsson2020b}, a first safe rule for SLOPE has been published~\citep{bao2020}. Yet, because of the non-separability of the penalty in SLOPE, this rule requires iterative screening during optimization, which means that predictors cannot be screened prior to fitting the model. This highlights the difficulty in developing screening rules for SLOPE.

Our main contribution in this paper is the presentation of a first heuristic screening rule for SLOPE based on the strong rule for the lasso. In doing so, we also introduce a novel formulation of the subdifferential for the sorted \(\ell_1\) norm. We then proceed to show that this rule is effective, rarely leads to violations, and offers performance gains comparable to the strong rule for the lasso.


\hypertarget{sec:notation}{%
  \subsection{Notation}\label{sec:notation}}

We use uppercase letters for matrices and
lowercase letters for vectors and scalars. \(\boldsymbol{1}\) and \(\boldsymbol{0}\)
denote vectors with all elements equal to 1 and 0 respectively, with
dimension inferred from context. We use \(\prec\) and \(\succ\) to denote
element-wise relational operators. We also let
\(\mathop{\mathrm{card}}\mathcal{A}\) denote the cardinality of set \(\mathcal{A}\) and define
\(\mathop{\mathrm{sign}}{x}\) to be the signum function with range \(\{-1,0,1\}\).
Furthermore, we define \(x_\downarrow\) to refer to a version of \(x\) sorted
in decreasing order and
the cumulative sum function for a vector \(x \in \mathbb{R}^n\) as
\(\mathop{\mathrm{cumsum}}(x) = [x_1, x_1 + x_2, \cdots,\sum_{i=1}^n x_i]^T.\)
We also let \({|i|}\) be the index operator of
\(y\in\mathbb{R}^p\) so that \(|y_{|i|}| = |y|_{(i)}\) for all \(i = 1,\dots,p\).
Finally, we allow a vector to be indexed with
an integer-valued set by eliminating those elements of this vector whose indices do not belong to the indexing set---for instance, if
\(\mathcal{A} = \{3,1\}\) and \(v = [v_1, v_2, v_3]^T\), then \(v_\mathcal{A} = [v_1, v_3]^T\).

\hypertarget{sec:screening-rules}{%
  \section{Theory}\label{sec:screening-rules}}

Proofs of the following theorem and propositions are provided in the supplementary material.

\hypertarget{sec:subdifferential}{%
  \subsection{The Subdifferential for SLOPE}\label{sec:subdifferential}}

The basis of the strong rule for \(\ell_1\)-regularized models is the subdifferential. By the same argument, we now turn to the subdifferential of SLOPE. The subdifferential for SLOPE has been derived previously as a characterization based on polytope mappings~\citep{bu2019a,tardivel2020}; here we present an alternative formulation that can be used as the basis of an efficient algorithm. First, however, let \(\mathcal{A}_i(\beta) \subseteq \{1,\dots,p\}\)
denote a set of indices for \(\beta \in \mathbb{R}^p\) such that
\begin{equation}
  \label{eq:cluster}
  \mathcal{A}_i(\beta) = \left\{ j \in \{1,\dots, p\}
  \mid \lvert \beta_i \rvert = \lvert \beta_j \rvert \right\}
\end{equation}
where
\(\mathcal{A}_i(\beta) \cap \mathcal{A}_l(\beta) = \varnothing\) if
\(l \not \in \mathcal{A}_i(\beta)\). To keep notation concise, we let
\(\mathcal{A}_i\) serve as a shorthand for \(\mathcal{A}_i(\beta)\).
In addition, we define the operator \(O : \mathbb{R}^p \rightarrow \mathbb{N}^p\),
which returns a permutation that rearranges its
argument in descending order by its absolute values
and \(R : \mathbb{R}^p \rightarrow \mathbb{N}^p\),
which returns the ranks of the absolute values in its argument.

\begin{example}
  If \(\beta = \{-3, 5, 3, 6\}\), then \(\mathcal{A}_1 = \{1, 3\}\),
  \(O(\beta) = \{4, 2, 1, 3\}\), and \(R(\beta) = \{3, 2, 4, 1\}\).
\end{example}

\begin{theorem}
  \label{thm:subgradient}
  The subdifferential \(\partial J(\beta;\lambda) \in \mathbb{R}^p\) is the
  set of all \(g \in \mathbb{R}^p\) such that
  \[
    g_{\mathcal{A}_i} =
    \left\{s \in \mathbb{R}^{\mathop{\mathrm{card}}{\mathcal{A}_i}} \bigm\vert
    \begin{cases}
      \mathop{\mathrm{cumsum}}(|s|_\downarrow - \lambda_{R(s)_{\mathcal{A}_i}}) \preceq \mathbf{0} & \text{if } \beta_{\mathcal{A}_i} = \mathbf{0}, \\
      \mathop{\mathrm{cumsum}}(|s|_\downarrow - \lambda_{R(s)_{\mathcal{A}_i}}) \preceq \mathbf{0}                                                  \\\quad \land \sum_{j \in \mathcal{A}_i}\left(|s_j| - \lambda_{R(s)_j}\right) = 0 \\
      \quad \land \sign \beta_{\mathcal{A}_i} = \sign s & \text{otherwise.}
    \end{cases}
    \right\}
  \]
\end{theorem}

\hypertarget{screening-rule-for-slope}{%
  \subsection{Screening Rule for SLOPE}\label{screening-rule-for-slope}}

\hypertarget{sparsity-pattern}{%
  \subsubsection{Sparsity Pattern}\label{sparsity-pattern}}

Recall that we are attempting to solve the following problem:
we know \(\hat\beta(\lambda^{(m)})\) and want to predict the support of
\(\hat\beta(\lambda^{(m+1)})\), where \(\lambda^{(m+1)} \preceq \lambda^{(m)}\).
The KKT stationarity criterion for SLOPE is
\begin{equation}
  \label{eq:kkt-stationarity}
  \boldsymbol{0} \in \nabla f(\beta) + \partial J(\beta;\lambda),
\end{equation}
where \(\partial J(\beta;\lambda)\) is the subdifferential
for SLOPE (\autoref{thm:subgradient}). This
means that if \(\nabla f(\hat\beta(\lambda^{(m+1)}))\) was available to us, we
could identify the support exactly. In \autoref{alg:sparsity-rule},
we present an algorithm to accomplish this in practice.

\begin{minipage}[t]{0.48\textwidth}
  \begin{algorithm}[H]
    \caption{\label{alg:sparsity-rule}}
    \begin{algorithmic}[1]
      \Require \(c \in \mathbb{R}^p\),
      \(\lambda \in \mathbb{R}^p\), where
      \(\lambda_1 \geq \cdots \geq \lambda_p \geq 0\).
      \State \(\mathcal{S}, \mathcal{B} \gets \varnothing\)
      \For{\(i \gets 1,\dots,p\)}
      \State \(\mathcal{B} \gets \mathcal{B} \cup \{i\}\)
      \If{\(\sum_{j \in \mathcal{B}}\big(c_j - \lambda_j\big) \geq 0\)}
      \State \(\mathcal{S} \gets \mathcal{S} \cup \mathcal{B}\)
      \State \(\mathcal{B} \gets \varnothing\)
      \EndIf
      \EndFor
      \State \Return \(\mathcal{S}\)
    \end{algorithmic}
  \end{algorithm}
\end{minipage}
\hfill
\begin{minipage}[t]{0.48\textwidth}
  \begin{algorithm}[H]
    \caption{Fast version of \autoref{alg:sparsity-rule}.\label{alg:fast-sparsity-rule}}
    \begin{algorithmic}[1]
      \Require \(c \in \mathbb{R}^p\), \(\lambda \in \mathbb{R}^p\), where
      \(\lambda_1 \geq \cdots \geq \lambda_p \geq 0\)
      \State \(i \gets 1\), \(k \gets 0\), \(s \gets 0\)
      \While{\(i + k \leq p\)}
      \State \(s \leftarrow s + c_{i+k} - \lambda_{i+k}\)
      \If{\(s \geq 0\)}
      \State \(k \gets k+i\)
      \State \(i \gets 1\)
      \State \(s \gets 0\)
      \Else
      \State \(i \gets i + 1\)
      \EndIf
      \EndWhile
      \State \Return \(k\)
    \end{algorithmic}
  \end{algorithm}
\end{minipage}


In \autoref{prop:rule-correctness}, we show that the result of
\autoref{alg:sparsity-rule} with \(c \coloneqq |\nabla f(\hat\beta(\lambda^{(m+1)}))|_\downarrow\)
and \(\lambda \coloneqq \lambda^{(m+1)}\) as input is certified to contain
the true support set of \(\hat\beta(\lambda^{(m+1)})\).

\begin{proposition}
  \label{prop:rule-correctness}
  Taking \(c \coloneqq \lvert\nabla f(\hat\beta(\lambda^{(m+1)}))\rvert_\downarrow\)
  and \(\lambda \coloneqq \lambda^{(m+1)}\) as input to
  \autoref{alg:sparsity-rule} returns a superset of the
  true support set of \(\hat\beta(\lambda^{(m+1)})\).
\end{proposition}

\begin{remark}
  In \autoref{alg:sparsity-rule}, we implicitly make use of the fact that
  the results are invariant to permutation changes within each
  cluster \(\mathcal{A}_i\) (as defined in \eqref{eq:cluster})---a fact that
  follows directly from the definition of the
  subdifferential~(\autoref{thm:subgradient}). In particular, this
  means that the indices for the set of inactive predictors will be
  ordered last in both \(|\hat\beta|_\downarrow\) and
  \(|\nabla f(\hat\beta)|_\downarrow\); that is, for all \(i,j \in \{1,2,\dots,p\}\)
  such that \(\hat\beta_i = 0\), \(\hat\beta_j \neq 0\),
  \[
    O(\nabla f(\hat\beta))_i > O(\nabla f(\hat\beta))_j \implies O(\hat\beta)_i > O(\hat\beta)_j,
  \]
  which allows us to determine the sparsity in \(\hat\beta\) via
  \(\nabla f(\hat\beta)\).
\end{remark}


\autoref{prop:rule-correctness} implies that \autoref{alg:sparsity-rule} may
lead to a conservative decision by potentially including some of
the support of inactive predictors
in the result, i.e. indices for which the corresponding coefficients are
in fact zero.
To see this, let \(\mathcal{U} = \{l,l+1,\dots,p\}\) be a set of inactive
predictors and take
\(c \coloneqq |\nabla f(\hat\beta(\lambda^{(m+1)}))|_\downarrow\).
For every \(k \in \mathcal{U}\), \(k \geq l\) for which
\(\sum_{i=l}^k(c_i - \lambda_i) = 0\),
\(\{l,l+1,\dots,k\}\) will be in the result
of \autoref{alg:sparsity-rule} in spite of being inactive.
This situation, however, occurs only when \(c\) is the
true gradient at the solution and for this reason is of little practical
importance.

Since the check in \autoref{alg:sparsity-rule} hinges only
on the last element of the cumulative sum at any given time, we need only to
store and update a single scalar instead of
the full cumulative sum vector. Using this fact, we can derive
a fast version of the rule (\autoref{alg:fast-sparsity-rule}), which
returns \(k\): the
predicted number of active predictors at the solution.\footnote{The active
  set is then retrieved by sub-setting the first \(k\) elements of the
  ordering permutation.}

Since we only have to take a single pass over the predictors, the
cost of the algorithm is linear in \(p\). To use the algorithm in practice,
however, we first need to compute the gradient at the previous solution
and sort it. Using least squares regression as an
example, this results in a complexity of \(\mathcal{O}(np + p\log p)\). To put
this into perspective, this is (slightly) lower than the cost of a single gradient
step if a first-order method is used to compute the SLOPE solution (since it
also requires evaluation of the proximal operator).

\hypertarget{sec:gradient-approximations}{%
  \subsubsection{Gradient Approximation}\label{sec:gradient-approximations}}

The validity of \autoref{alg:sparsity-rule} requires
\(\nabla f(\hat\beta(\lambda^{(m+1)}))\) to be available, which of course
is not the case. Assume, however, that we are given a reasonably accurate
surrogate of the gradient vector and suppose that we
substitute this estimate for \(\nabla f(\hat\beta(\lambda^{(m+1)}))\)
in \autoref{alg:sparsity-rule}. Intuitively, this should yield us an
estimate of the active set---the better the approximation, the more closely
this screened set should resemble the active set. For the sequel,
let \(\mathcal{S}\) and \(\mathcal{T}\) be the screened and active set
respectively.

An obvious consequence of using our approximation is that we
run the risk of picking \(\mathcal{S} \not\supseteq \mathcal{T}\),
which we then naturally must safeguard against. Fortunately, doing so
requires only a KKT stationarity check---whenever the check fails, we relax
\(\mathcal{S}\) and refit. If such failures are rare,
it is not hard to imagine that the benefits of tackling
the reduced problem
might outweigh the costs of these occasional failures.

Based on this argument, we are now ready to
state the strong rule for SLOPE, which is a
natural extension of the strong rule for the lasso~\citep{tibshirani2012}. Let
\(\mathcal{S}\) be the output from running
\autoref{alg:sparsity-rule} with
\[
  c \coloneqq \big(|\nabla f(\hat\beta(\lambda^{(m)}))| + \lambda^{(m)} - \lambda^{(m+1)}\big)_\downarrow,\qquad \lambda \coloneqq \lambda^{(m+1)}
\]
as input. The strong rule for SLOPE
then discards all predictors corresponding to \(\mathcal{S}^\mathsf{c}\).

\begin{proposition}
  \label{prop:strong-rule}
  Let \(c_j(\lambda) = (\nabla f(\hat\beta(\lambda)))_{|j|}\).
  If \(|c_j'(\lambda)| \leq 1\) for all \(j=1,2,\dots,p\) and
  \(O(c(\lambda^{(m+1)})) = O(c(\lambda^{(m)}))\) (see \autoref{sec:subdifferential}
  for the definition of \(O\)), the strong rule for SLOPE returns a superset of the true support set.
\end{proposition}

Except for the assumption on fixed ordering permutation,
the proof for \autoref{prop:strong-rule} is comparable to
the proof of the strong rule for the lasso~\citep{tibshirani2012}. The
bound appearing in the proposition, \(|c_j'(\lambda)| \leq 1\), is
referred to as the \emph{unit slope bound}, which
results in the following rule for the lasso: discard the \(j\)th predictor if
\[
  \big|\nabla f(\beta(\lambda^{(m)}))_j\big| \leq 2\lambda^{(m+1)}_j - \lambda^{(m)}_j.
\]
In \autoref{prop:strong-generalization}, we formalize the connection
between the strong rule for SLOPE and lasso.

\begin{proposition}
  \label{prop:strong-generalization}
  The strong rule for SLOPE is a generalization of the strong rule for
  the lasso; that is, when \(\lambda_j=\lambda_i\) for all \(i,j\in \{1,\dots,p\}\),
  the two rules always produce the same screened set.
\end{proposition}

Finally, note that a non-sequential (basic) version of this rule is obtained by simply using the gradient for the null model as the basis for the approximation together with the penalty sequence corresponding to the point at which the first predictor enters the model (see \autoref{setup}).

\hypertarget{violations-of-the-rule}{%
  \subsubsection{Violations of the Rule}\label{violations-of-the-rule}}

Violations of the strong rule for SLOPE occur only when the unit slope bound fails, which may happen for both inactive and active predictors---in the latter case, this can occur when the clustering or the ordering permutation changes for these predictors. This means that the conditions under which violations may arise for the strong rule for SLOPE differ from those corresponding to the strong rule for the lasso~\citep{tibshirani2012}.


To safeguard against violations, we check the KKT conditions after each fit and add violating predictors to the screened set, refit, and repeat the checks until there are no violations. In \autoref{sec:violations-simulated}, we will study the prevalence of violations in simulated experiments.

\hypertarget{sec:algorithms}{%
  \subsubsection{Algorithms}\label{sec:algorithms}}

\citet{tibshirani2012} considered two algorithms using the
strong rule for the lasso. In this paper, we consider
two algorithms that are analogous except in one regard. First, however, let
\(\mathcal{S}(\lambda)\) be the strong set, i.e. the set obtained by application
of the strong rule for SLOPE, and \(\mathcal{T}(\lambda)\) the active set.
Both algorithms begin with a set \(\mathcal{E}\) of predictors, fit
the model to this set, and then either expand this set, refit and repeat, or
stop.

In the \emph{strong set} algorithm (see supplementary material for details)
we initialize \(\mathcal{E}\) with the
union of the strong set and the set
of predictors active at the previous step on the regularization path.
We then fit the model
and check for KKT violations in the full set of predictors, expanding
\(\mathcal{E}\) to include any predictors for which violations occur and
repeat until there are no violations.

In the \emph{previous set} algorithm (see supplementary material for details)
we initialize \(\mathcal{E}\) with only
the set of previously active predictors, fit, and check the KKT conditions
against the strong rule set. If there are violations in the strong set,
the corresponding predictors are added to \(\mathcal{E}\) and the model
is refit.
Only when there are no violations in the strong set do we check the
KKT conditions in the full set. This procedure is repeated until there are
no violations in the full set.

These two algorithms differ from the strong and working set algorithms from
\citet{tibshirani2012} in that we use only the set of previously active
predictors rather than the set of predictors that have
been active at any previous step on the path.

\hypertarget{sec:experiments}{%
  \section{Experiments}\label{sec:experiments}}

In this section we present simulations that examine the effects of
applying the screening rules. The problems here reflect our focus on
problems in the \(p \gg n\) domain, but we will also briefly
consider the reverse in order to examine the potential overhead
of the rules when \(n > p\).


\hypertarget{setup}{%
  \subsection{Setup}\label{setup}}

Unless stated otherwise, we will use
the strong set algorithm with the
strong set computed using \autoref{alg:fast-sparsity-rule}.
Unless stated otherwise, we normalize the predictors such that \(\bar x_j = 0\)
and \(\lVert x_j \rVert_2 = 1\) for \(j = 1,\dots,p\). In addition, we center
the response vector such that \(\bar y = 0\) when \(f(\beta)\) is the least squares objective.

We use the \emph{Benjamini--Hochberg}~(BH) method~\citep{bogdan2015} for computing the
sequence, which sets \(\lambda_i^\mathrm{BH} = \Phi^{-1}\big(1 - qi/(2p)\big)\) for \(i= 1,2,\dots,p\),
where \(\Phi^{-1}\) is the probit function.\footnote{\citet{bogdan2015} also presented
  a method called the \emph{Gaussian} sequence that is a modification of the
  BH method, but it is not appropriate for our problems since it reduces to the lasso
  in the \(p \gg n\) context.}
To construct the regularization path, we parameterize the sorted \(\ell_1\)
penalty as \(J(\beta; \lambda,\sigma) = \sigma \sum_{j=1}^p|\beta|_{(j)}\lambda_j\),
with \(\sigma^{(1)} > \sigma^{(2)} > \cdots > \sigma^{(l)} > 0\).
We pick \(\sigma^{(1)}\) corresponding to
the point at which the first predictor enters the model, which
corresponds to maximizing \(\sigma \in \mathbb{R}\) subject to
\(\mathop{\mathrm{cumsum}}(\nabla f(\boldsymbol{0})_\downarrow - \sigma\lambda) \preceq0\),
which is given explicitly as \[\sigma^{(1)} = \max (\mathop{\mathrm{cumsum}}(\nabla f(\boldsymbol{0})_\downarrow) \oslash \mathop{\mathrm{cumsum}}(\lambda)),\]
where \(\oslash\) is the Hadamard (element-wise) division operator. We choose
\(\sigma^{(l)}\) to be \(t\sigma^{(1)}\) with \(t= 10^{-2}\) if \(n < p\) and \(10^{-4}\)
otherwise. Unless stated otherwise, we employ a regularization path of
\(l = 100\) \(\lambda\) sequences but stop this path prematurely if
1) the number of unique coefficient magnitudes exceed the number of observations, 2) the
fractional change in deviance from one step to another is less than \(10^{-5}\), or 3) if the
fraction of deviance explained exceeds \(0.995\).

Throughout the paper we use version 0.2.1 of the R
package \pkg{SLOPE}~\citep{larsson2020}, which uses
the accelerated proximal gradient algorithm FISTA~\citep{beck2009} to
estimate all models; convergence is obtained when the duality gap as a fraction of the primal and the relative level of infeasibility~\citep{bogdan2015a} are lower than \(10^{-5}\) and \(10^{-3}\) respectively.  All simulations
were run on a dedicated high-performance computing cluster and the
code for the simulations is available in the supplementary material and at \url{https://github.com/jolars/slope-screening-code/}.

\hypertarget{sec:simulated-data}{%
  \subsection{Simulated Data}\label{sec:simulated-data}}

Let \(X \in \mathbb{R}^{n \times p}\), \(\beta \in \mathbb{R}^{p \times m}\),
and \(y \in \mathbb{R}^n\). We take
\[
  y_i = x_i^T\beta + \varepsilon_i, \qquad i = 1,2,\dots,n,
\]
where \(\varepsilon_i\) are sampled from independently and identically
distributed standard normal variables. \(X\) is generated such that each
row is sampled independently and identically from
a multivariate normal distribution \(\mathcal{N}(\boldsymbol{0}, \Sigma)\).
From here on
out, we also let \(k\) denote the cardinality of the non-zero support set of the
true coefficients, that is,
\(k = \mathop{\mathrm{card}}\{i \in \mathbb{N}^p \mid \beta_i \neq 0\}\).

\hypertarget{sec:efficiency-simulated}{%
  \subsubsection{Efficiency}\label{sec:efficiency-simulated}}

We begin by studying the efficiency of the strong rule for SLOPE on problems with varying levels of correlation \(\rho\). Here, we let \(n=200\), \(p=5000\), and \(\Sigma_{ij} = 1\) if \(i=j\) and \(\rho\) otherwise.
We take \(k=p/4\) and generate \(\beta_i\) for \(i=1,\dots,k\)
from \(\mathcal{N}(0,1)\). We then fit a least squares regression model regularized
with the sorted \(\ell_1\) norm to this data and screen the predictors
with the strong rule for SLOPE. Here
we set \(q=0.005\) in the construction of the BH
sequence.

\begin{figure}[hbt]
  \centering
  \includegraphics{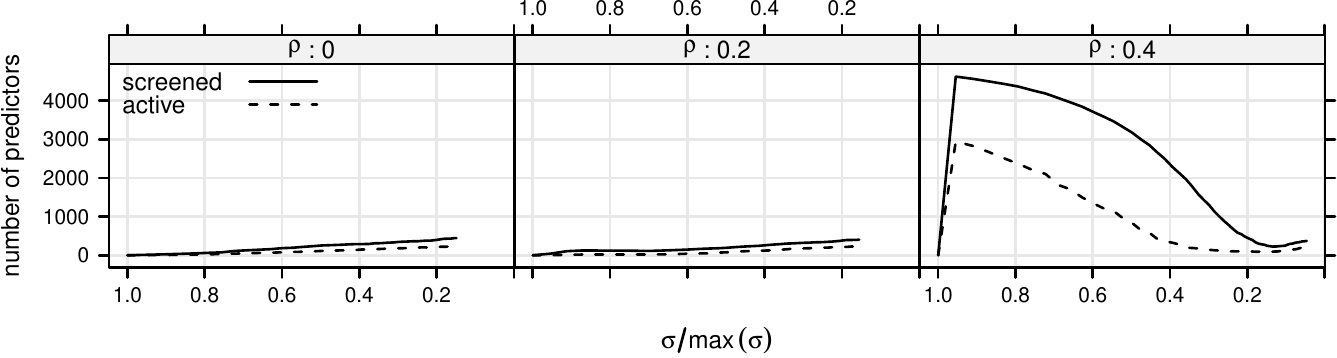}
  \caption{Number of screened and active predictors for sorted \(\ell_1\)-regularized least squares regression using no screening or the strong rule for SLOPE.}\label{fig:gaussian-correlated}
\end{figure}

The size of the screened set is clearly small next to the full set~ (\autoref{fig:gaussian-correlated}). Note, however, that the presence of strong correlation among the predictors both means that there is less to be gained by screening since many more predictors are active at the start of the path, as well as makes the rule more conserative. No violations of the rule were observed in these simulations.

\hypertarget{sec:violations-simulated}{%
  \subsubsection{Violations}\label{sec:violations-simulated}}

To examine the number of violations of the rule, we generate a number of data sets with
\(n=100\), \(p \in \{20, 50, 100, 500, 1000\}\), and \(\rho = 0.5\).
We then fit a full path of 100 \(\lambda\) sequences across 100 iterations, averaging the results.
(Here we disable the rules for prematurely aborting the path described
at the start of this section.)
We sample the first fourth of the elements of
\(\beta\) from \(\{-2,2\}\) and set the rest to zero.

\begin{figure}[hbt]
  \centering
  \includegraphics{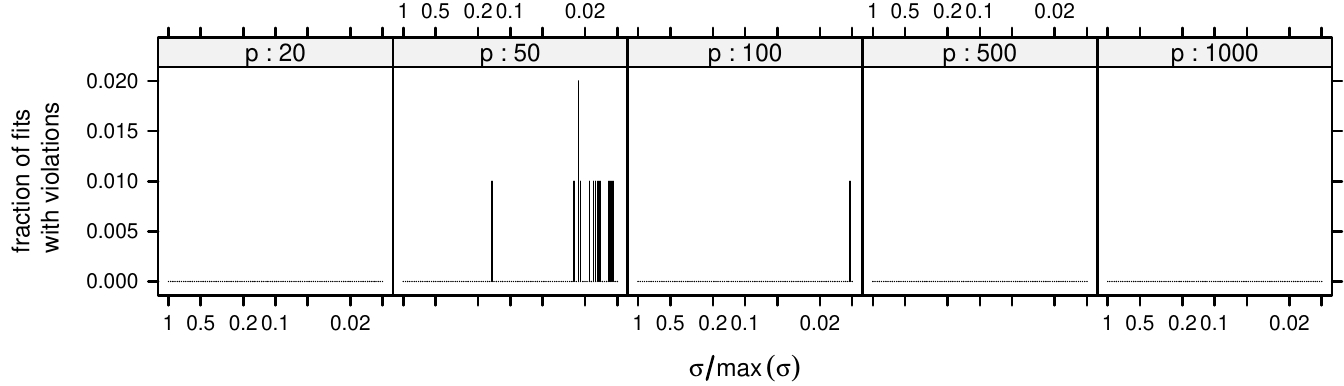}
  \caption{Fraction of model fits resulting in violations of the strong rule for sorted \(\ell_1\)-regularized least squares regression.}\label{fig:gaussian-correlated-violations}
\end{figure}

Violations appear to be
rare in this setting and occur only for the lower range of \(p\) values~(\autoref{fig:gaussian-correlated-violations}). For \(p=100\), for instance,
we would at an average need to estimate roughly 100 paths for
this type of design to encounter a single violation. Given that
a complete path consists of 100 steps and that the warm start after
the violation is likely a good initialization, this can be considered a marginal
cost.

\hypertarget{sec:performance-simulated}{%
  \subsubsection{Performance}\label{sec:performance-simulated}}

In this section, we study the performance of the screening rule for
sorted \(\ell_1\)-penalized
least squares, logistic, multinomial, and Poisson regression.

We now take \(p=20,000\), \(n=200\), and \(k=20\). To construct \(X\), we let
\(X_1,X_2,\dots,X_p\) be random variables distributed according to
\[
  X_1 \sim \mathcal{N}(\boldsymbol{0}, I),\qquad X_j \sim \mathcal{N}(\rho X_{j-1},I) \quad\text{for } j=2,3,\dots,p,
\]
and sample the \(j\)th column in \(X\) from \(X_j\) for \(j=1,2,\dots,p\).

\begin{figure}[hbtp]
  \vspace{-5ex}
  \includegraphics{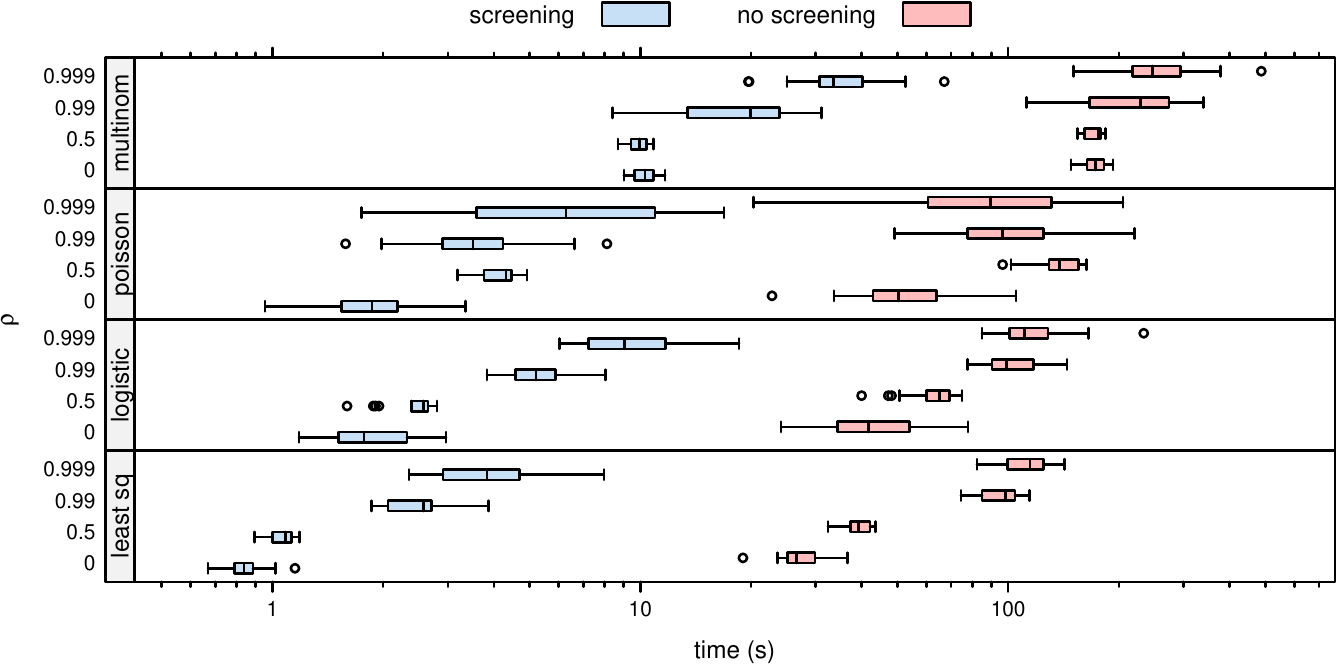}
  \caption{Time taken to fit SLOPE with or without the strong screening rule for randomly generated data.}\label{fig:performance-simulated-data}
\end{figure}

For least squares and logistic regression data we
sample the first \(k=20\) elements of \(\beta\) without replacement from
\(\{1,2,\dots,20\}\). Then we let \(y = X\beta + \varepsilon\)
for least squares regression
and \(y = \mathop{\mathrm{sign}}{(X\beta + \varepsilon)}\) for logistic regression, in both
cases taking \(\varepsilon \sim \mathcal{N}(\boldsymbol{0}, 20I)\).
For Poisson regression, we generate
\(\beta\) by taking random samples without replacement from
\(\{\frac{1}{40}, \frac{2}{40}, \dots, \frac{20}{40}\}\) for its first 20
elements. Then we sample \(y_i\) from
\(\operatorname{Poisson}\big(\exp((X\beta)_i)\big)\) for \(i=1,2,\dots,n\).
For multinomial regression, we start by
taking \(\beta \in \mathbb{R}^{p\times 3}\), initializing all elements to
zero. Then, for each row in \(\beta\) we
take a random sample from \(\{1,2,\dots,20\}\) without replacement and insert
it at random into one of the elements of that row.
Then we sample \(y_i\) randomly from
\(\mathrm{Categorical}(3, p_i)\) for \(i=1,2,\dots,n\),
where
\[
  p_{i,l} = \frac{\exp\big((X\beta)_{i,l}\big)}{\sum_{l=1}^3 \exp\big((X\beta)_{i,l}\big)}.
\]
The benchmarks reveal a strong effect on account of the
screening rule through the range of
models used (\autoref{fig:performance-simulated-data}), leading
to a substantial reduction in run time. As an example,
the run time for fitting logistic regression when \(\rho = 0.5\)
decreases from roughly 70 to 5 seconds when the screening rule is used.

We finish this section with an examination of two types of algorithms
outlined in \autoref{sec:algorithms}: the strong set
and previous set algorithm. In
\autoref{fig:gaussian-correlated} we observed that the strong rule
is conservative when correlation is high among predictors, which
indicates
that the previous set algorithm might yield an improvement over
the strong set algorithm.

In order to examine this, we
conduct a simulation in which we vary the strength of correlation
between predictors as well as the parameter \(q\) in the construction of the
BH regularization sequence. Motivation for varying the latter comes from the
relationship between coefficient clustering and
the intervals in the regularization
sequence---higher values of \(q\) cause larger gaps in the sequence, which
in turn leads to more clustering among predictors. This clustering, in turn,
is strongest at the start of the path when regularization is strong.

For large enough \(q\) and \(\rho\), this behavior in
fact occasionally causes almost all predictors to
enter the model at the second step on the path. As an example, using
when \(\rho = 0.6\) and fitting with \(q=10^{-2}\) and \(10^{-4}\) leads to 2215
and 8 nonzero coefficients respectively at the second step in one simulation.

Here, we
let \(n=200\), \(p=5000\), \(k=50\), and \(\rho \in \{0,0.1,0.2,\dots,0.8\}\). The
data generation process corresponds to the setup at the start of this section for
least squares regression data except for the covariance structure of \(X\), which is equal to that in
\autoref{sec:efficiency-simulated}.
We sample the non-zero entries in \(\beta\) independently
from a random variable \(U \sim \mathcal{N}(0,1)\).

\begin{wrapfigure}[13]{r}{0.3\linewidth}
  \vspace{-11ex}
  \centering
  \includegraphics{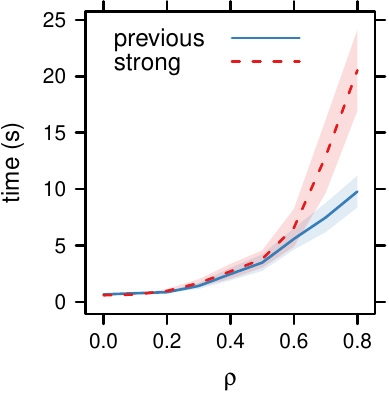}
  \caption{Time taken to fit a regularization path of SLOPE for least squares regression using either the strong or previous set algorithm.}\label{fig:performance-alg}
\end{wrapfigure}

The two algorithms perform similarly for
\(\rho \leq 0.6\) (\autoref{fig:performance-alg}). For larger \(\rho\), the
previous set strategy evidently outperforms the strong set strategy. This
result is not surprising: consider \autoref{fig:gaussian-correlated},
for instance, which shows that the behavior of the regularization path under
strong correlation makes the previous set strategy particularly effective
in this context.

\hypertarget{sec:real-data}{%
  \subsection{Real Data}\label{sec:real-data}}

\hypertarget{sec:efficiency-real}{%
  \subsubsection{Efficiency and Violations}\label{sec:efficiency-real}}

We examine efficiency and violations for four real data sets: \emph{arcene}, \emph{dorothea}, \emph{gisette}, and \emph{golub}, which are the same data sets that were examined in \citet{tibshirani2012}. The first three originate from \citet{guyon2004} and were originally collected from the UCI (University of California Irvine) Machine Learning Repository~\citep{dua2019}, whereas the last data set, \emph{golub}, was originally published in \citet{golub1999}. All of the data sets were collected from \url{http://statweb.stanford.edu/~tibs/strong/realdata/} and feature a response \(y \in \{0,1\}\). We fit both least squares and logistic regression models to the data sets and examine the effect of the level of coarseness in the path by varying the length of the path (\(l = 20,50,100\)).

\begin{figure}[hbt]
  \centering
  \includegraphics{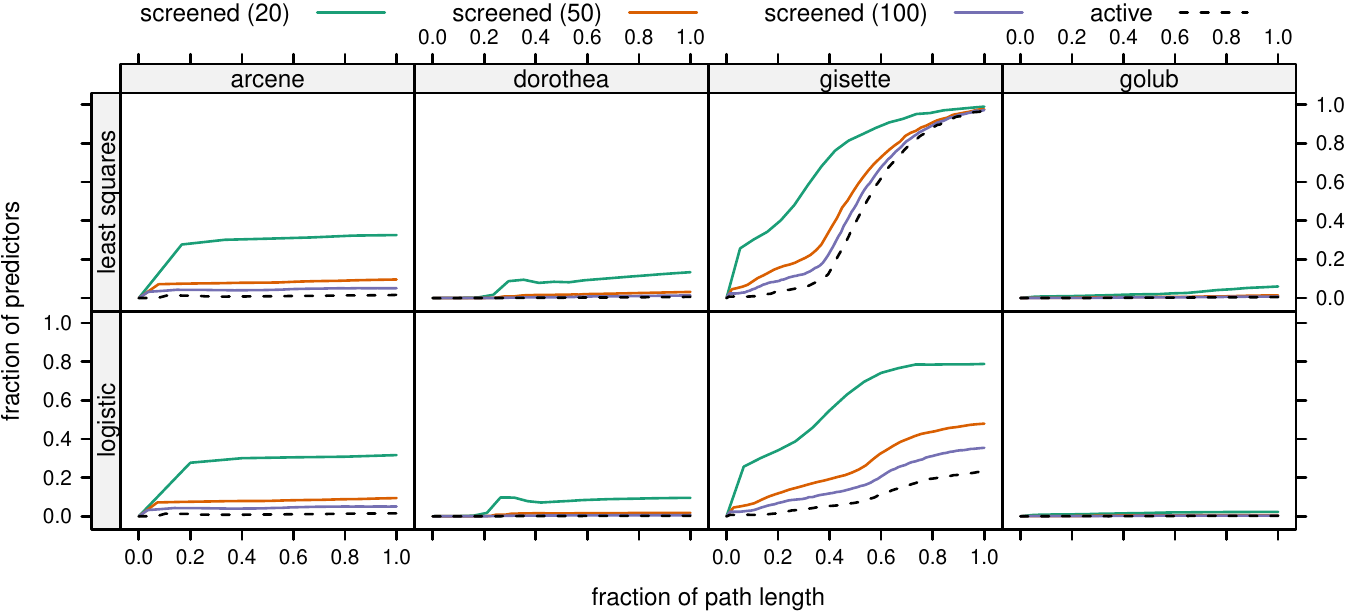}
  \caption{Proportion of predictors included in the model by the strong screening rule as a proportion of the total number of active predictors in the model for a path of \(\lambda\) sequences. Three types of paths have are examined, using path lengths of 20, 50, and 100.}\label{fig:efficiency-real-data}
\end{figure}

There were no violations in any of the fits. The screening rule offers substantial reductions in problem size  (\autoref{fig:efficiency-real-data}), particularly for the path length of 100, for which the size of the screened set of predictors ranges from roughly 1.5--4 times the size of the active set.

\hypertarget{sec:performance-real}{%
  \subsubsection{Performance}\label{sec:performance-real}}

In this section, we introduce three new data sets: \emph{e2006-tfidf}~\citep{frandi2015}, \emph{physician}~\citep{deb1997}, and \emph{news20}~\citep{lang1995a}. \emph{e2006-tfidf} was collected from \citet{frandi2015}, \emph{news20} from \url{https://www.csie.ntu.edu.tw/~cjlin/libsvmtools/datasets}~\citep{chang2011}, and \emph{physician} from \url{https://www.jstatsoft.org/article/view/v027i08}~\citep{zeileis2008}. We use the test set for \emph{e2006-tfidf} and a subset of 1000 observations from the training set for \emph{news20}.

\begin{table}
  \caption{\label{tab:perf-real-data}Benchmarks measuring wall-clock time for four data sets fit with different models using either the strong screening rule or no rule.}
  \centering
  \small
  \begin{tabular}[t]{llrrrr}
    \toprule
    \multicolumn{1}{c}{} & \multicolumn{1}{c}{} & \multicolumn{1}{c}{} & \multicolumn{1}{c}{} & \multicolumn{2}{c}{time (s)}             \\
    \cmidrule(l{3pt}r{3pt}){5-6}
    dataset              & model                & $n$                  & $p$                  & no screening                 & screening \\
    \midrule
    dorothea             & logistic             & 800                  & 88119                & 914                          & 14        \\
    e2006-tfidf          & least squares        & 3308                 & 150358               & 43353                        & 4944      \\
    news20               & multinomial          & 1000                 & 62061                & 5485                         & 517       \\
    physician            & poisson              & 4406                 & 25                   & 34                           & 34        \\
    \bottomrule
  \end{tabular}
\end{table}

In \autoref{tab:perf-real-data}, we summarize the results from
fitting sorted \(\ell_1\)-regularized least squares, logistic, Poisson, and multinomial
regression to the four data sets.
Once again, we see that the screening rule improves performance in the
high-dimensional regime and presents no noticeable drawback even when \(n > p\).

\hypertarget{conclusions}{%
  \section{Conclusions}\label{conclusions}}

In this paper, we have developed a heuristic predictor screening rule for SLOPE
and shown that it is a generalization of the strong rule for the lasso.
We have demonstrated that it offers dramatic improvements in the \(p \gg n\)
regime, often reducing the time required to fit the full regularization
path for SLOPE by orders of
magnitude, as well as imposing little-to-no cost when \(p < n\).
At the time of this publication, an efficient implementation of
the screening rule is available in the R package \pkg{SLOPE}~\citep{larsson2020}.

The performance of the rule is demonstrably weaker when predictors in the
design matrix are heavily correlated. This issue may be mitigated by the use of the previous set strategy that we have investigated here; part of the problem, however, is related to the clustering behavior that SLOPE exhibits: large portions of the total
number of predictors often enter the model in a few clusters when
regularization is strong. A possible avenue for future research might therefore be to investigate if screening rules for this clustering behavior might be
developed and utilized to further
enhance performance in estimating SLOPE.

\hypertarget{broader-impact}{%
  \section*{Broader Impact}\label{broader-impact}}
\addcontentsline{toc}{section}{Broader impact}

The predictor screening rules introduced in this article
allow for a substantial improvement of the speed of SLOPE. This
facilitates application of SLOPE to the identification of
important predictors in huge data bases, such as collections of whole genome genotypes
in Genome Wide Association Studies. It also paves the way for the
implementation of cross-validation techniques and improved efficiency
of the Adaptive Bayesian version SLOPE (ABSLOPE~\citep{jiang2019}), which requires multiple iterations of the
SLOPE algorithm. Adaptive SLOPE bridges Bayesian and the frequentist
methodology and enables good predictive models with FDR control in
the presence of many hyper-parameters or missing data. Thus it
addresses the problem of false discoveries and
lack of replicability in a variety of important
problems, including medical and genetic studies.

In general, the improved efficiency resulting from the predictor screening rules
will make the SLOPE family of models (SLOPE~\citep{bogdan2015}, grpSLOPE~\citep{brzyski2018}, and ABSLOPE) accessible to a broader audience, enabling researchers
and other parties to fit SLOPE models with improved efficiency. The time required
to apply these models will be reduced and, in some cases, data sets that were otherwise too
large to be analyzed without access to dedicated high-performance computing clusters
can be tackled even with modest computational means.

We can think of no way by which these screening rules may put anyone at
disadvantage. The methods we outline here do not in any way affect the
model itself (other than boosting its performance) and can therefore only be of
benefit. For the same reason, we do not believe that the strong rules for SLOPE
introduces any ethical issues, biases, or negative societal consequences. In contrast,
it is in fact possible that the reverse is true given that SLOPE serves as an alternative
to, for instance, the lasso, and has superior
model selection properties~\citep{figueiredo2016,jiang2019} and
lower bias~\citep{jiang2019}.

\begin{ack}
  We would like to thank Patrick Tardivel (Institut de Math\'ematiques de Bourgogne) and Yvette Baurne (Department of Statistics, Lund University)
  for their insightful comments on the paper. In particular, we would like to 
  thank Patrick Tardivel for noticing an error in the formulation of
  \autoref{thm:subgradient} in a previous version of this paper.
  
  The computations were enabled by resources provided by the Swedish National Infrastructure for Computing (SNIC) at Lunarc partially funded by the Swedish Research Council through grant agreement no. 2018-05973.

  The research of Ma{\l}gorzata Bogdan was supported by the grant of the Polish National Center of Science no. 2016/23/B/ST1/00454. The research of Jonas Wallin was supported by the Swedish Research Council, grant no. 2018-01726.  These entities, however, have in no way influenced the work on this paper.
\end{ack}

\bibliography{references}

\appendix

\hypertarget{proofs}{%
  \section{Proofs}\label{proofs}}

\hypertarget{proof-theorem1}{%
  \subsection{Proof of Theorem 1}\label{proof-theorem1}}

By definition, the subdifferential \(\partial J(\beta;\lambda)\) is the
set of all \(g \in \mathbb{R}^p\) such that
\begin{equation}
  \label{eq:subgrad}
  J(y;\lambda) \geq J(\beta;\lambda) + g^T(y-\beta)
  = \sum_{j=1}^p |\beta|_{(j)}\lambda_j + g^T(y-\beta),
\end{equation}
for all \(y \in \mathbb{R}^p\).

Assume that we have \(K\) clusters
\(\mathcal{A}_1,\mathcal{A}_2,\dots,\mathcal{A}_K\)
(as defined per Equation 2 (main article)) and that \(\beta = |\beta|_\downarrow\),
which means we can rewrite \eqref{eq:subgrad} as
\[
  \begin{aligned}
    0 & \geq J(\beta;\lambda) - J(y;\lambda) + g^T(y - \beta)                                                     \\
      & = \sum_{i \in \mathcal{A}_1}(\lambda_i|\beta|_{(i)} - g_i\beta_i -\lambda_i|y|_{(i)} + g_iy_i) + \dots    \\
      & \phantom{=}+\sum_{i \in \mathcal{A}_K}(\lambda_i|\beta|_{(i)} - g_i\beta_i -\lambda_i|y|_{(i)} + g_iy_i).
  \end{aligned}
\]
Notice that we must have
\(\sum_{i \in \mathcal{A}_j}(\lambda_i|\beta|_{(i)} - g_i\beta_i -\lambda_i|y|_{(i)} + g_iy_i) \leq 0\)
for all \(j \in \{1,2,\dots,K\}\) since otherwise the inequality breaks by
selecting \(y_i = \beta_i\) for \(i \in \mathcal{A}^c_j\). This means that it is
sufficient to restrict attention to a single set as well as
take this to be the set \(\mathcal{A}_i = \{1,\dots,p\}\).

\begin{case}[\(\beta = \mathbf{0}\)]
  In this case \eqref{eq:subgrad} reduces
  to \(J(y;\lambda) \geq g^Ty\). Now take a \(c \in \mathcal{Z}\) where
  \begin{equation}
    \label{eq:z-set}
    \mathcal{Z} = \left\{s \in \mathbb{R}^p\bigm\vert \mathop{\mathrm{cumsum}}(|s|_\downarrow - \lambda) \preceq \mathbf{0}\right\}
  \end{equation}
  and assume that \(|c_1| \geq \cdots \geq |c_p|\) without loss of generality.

  Clearly, \(J(y;\lambda) \geq c^Ty\) holds if and only if
  \(J(y^*;\lambda) - c^Ty^* \geq 0\) where
  \[
    y^* = \mathop{\mathrm{arg\,min}}_y \left\{ J(y;\lambda) - c^Ty\right\}.
  \]
  Now, since \(J(y;\lambda)\) is invariant to changes in signs and permutation of
  \(y\), it follows from the rearrangement inequality~\citep[Theorem 368]{hardy1952} that \(|y|^*_1 \geq \cdots \geq |y|^*_p\).
  This permits us to formulate the following equivalent problem:
  \[
    \begin{aligned}
       & \text{minimize}   &  & y^T(\mathop{\mathrm{sign}}(y)\odot\lambda - c)         \\
       & \text{subject to} &  & \mathop{\mathrm{sign}}(y) = \mathop{\mathrm{sign}}(c), \\
       &                   &  & |y_1| \geq \cdots \geq |y_p|.                          \\
    \end{aligned}
  \]

  To minimize the objective \(y^T(\mathop{\mathrm{sign}}(y)\odot\lambda - |c|) = |y|^T(\lambda - |c|)\),
  recognize first that we must have \(y^*_1 = y_2^*\)
  since \(c \in \mathcal{Z}\), which implies \(\lambda_1 - |c_1| \geq 0\). Likewise, \(y_2^*(\lambda_1 - |c_1|) + y_2^*(\lambda_2 - |c_2|) \geq 0\)
  since \(\lambda_1 + \lambda_2 - (|c_1| + |c_2|) \geq 0\), which leads us to conclude that
  \(y^*_2 = y^*_3\). Then, proceeding inductively,
  it is easy to see that \(y_p^* \sum_{i=1}^p(\lambda_i - |c_i|) \geq 0\),
  which implies \(y^*_1 = \cdots = y^*_p = 0\). At this point, we have
  shown that \(c \in \mathcal{Z} \implies c \in \partial J(\beta;\lambda)\).

  For the next part note that \(g \in \mathcal{Z}\) is equivalent to requiring \(|g|_{(1)} \leq \lambda_1\) and
  \begin{equation}
    \label{eq:z-equivalent}
    |g|_{(i)} \leq \sum_{j=1}^i \lambda_j - \sum_{j=2}^{i}|g|_{(j)}, \qquad i=1,\dots,p.
  \end{equation}
  Now assume that there is a
  \(c\) such that \(c \in \partial J(\beta;\lambda)\) and \(c \notin \mathcal{Z}\).
  Then there exists an \(\varepsilon > 0\) and \(i \in \{1,2,\dots,p\}\) such that
  \[
    |c|_{(i)} \leq \sum_{j=1}^i \lambda_j - \sum_{j=2}^{i}|c|_{(j)} + \varepsilon, \qquad i=1,\dots,p.
  \]
  Yet if
  \(c = [\lambda_1, \dots, \lambda_{i-1}, \lambda_i + \varepsilon, \lambda_{i+1}, \dots, \lambda_p]^T\)
  then \eqref{eq:subgrad} breaks for \(y = \boldsymbol{1}\), which implies that
  \(c \notin \mathcal{Z} \implies c \notin \partial J(\beta;\lambda)\).
\end{case}
\begin{case}[\(\beta \neq \mathbf{0}\)]
  Now let \(|\beta_i| \coloneqq \alpha\) for all
  \(i = 1,\dots,p\), since by construction all \(\beta\) are equal in absolute
  value. Now \eqref{eq:subgrad} reduces to
  \begin{equation}
    \label{eq:nonzero-inequality}
    \begin{aligned}
      J(y;\lambda) & \geq J(\beta;\lambda) - g^T\beta + g^Ty                                                        \\
                   & = \sum_{i=1}^p \lambda_i \alpha  - \sum_{i=1}^pg_i\mathop{\mathrm{sign}}(\beta_i)\alpha + g^Ty \\
                   & = \alpha \sum_{i=1}^p\left( \lambda_i - g_i\mathop{\mathrm{sign}}(\beta_i)\right) + g^Ty.
    \end{aligned}
  \end{equation}
  The first term on the right-hand side of the last equality must be zero
  since otherwise the inequality breaks for \(y = \boldsymbol{0}\). In addition,
  it must also hold that \(\mathop{\mathrm{sign}}(\beta_i) =\mathop{\mathrm{sign}}(g_i)\) for all \(i\)
  such that \(|\beta_i| > 0\).
  To show this, suppose the opposite is true, that is, there exists at least
  one \(j\) such that \(\mathop{\mathrm{sign}}(g_j) \neq \mathop{\mathrm{sign}}(\beta_j)\). But then if
  we take \(y_j = \alpha\mathop{\mathrm{sign}}(g_j)\) and \(y_i = -\alpha\mathop{\mathrm{sign}}(g_i)\),
  \eqref{eq:nonzero-inequality}
  is violated, which proves the statement by contradiction.

  Taken together, this means that
  we have \(g \in \mathcal{H}\) where
  \[
    \mathcal{H} = \left\{s \in \mathbb{R}^p \mid \sum_{j = 1}^p \left(|s_j| - \lambda_j\right) = 0 .\right\}
  \]
  We are now left
  with \(J(y;\lambda) \geq g^Ty\), but this is exactly
  the setting from case one. Direct application of the reasoning from that
  part shows that we must have \(g \in \mathcal{Z}\).
  Connecting the dots, we finally conclude that
  \(c \in \mathcal{Z} \cap \mathcal{H} \implies c \in \partial J(\beta;\lambda)\).
\end{case}

\hypertarget{proof-proposition1}{%
  \subsection{Proof of Proposition 1}\label{proof-proposition1}}

Suppose that we have \(\mathcal{B} \neq \varnothing\) after running
Algorithm 1 (main article). In this case we have
\[
  \mathop{\mathrm{cumsum}}(c_\mathcal{B}-\lambda_\mathcal{B}) =
  \mathop{\mathrm{cumsum}}\bigg(\Big(\big|\nabla f(\hat\beta(\lambda^{(m+1)}))\big|_\downarrow\Big)_{\mathcal{B}}
  - \lambda^{(m+1)}_\mathcal{B}\bigg) \prec \boldsymbol{0},
\]
which
implies via Theorem 1 (main article) and Equation 3 (main article) that
all predictors in \(\mathcal{B}\) must be inactive and that \(\mathcal{S}\)
contains the true support set.

\hypertarget{proof-proposition2}{%
  \subsection{Proof of Proposition 2}\label{proof-proposition2}}

We need to show that the strong rule approximation does not violate the
inequality on the fourth line in Algorithm 1 (main article).
Since \(\mathop{\mathrm{cumsum}}(y) \succeq \mathop{\mathrm{cumsum}}(x)\) for all \(x, y \in \mathbb{R}^p\)
if and only iff \(y \succeq x\), it suffices to show that
\[
  \lvert c_j(\lambda^{(m)})\rvert + \lambda^{(m)}_j - \lambda_j^{(m+1)} \geq  |c_j(\lambda^{(m+1)})|
\]
for all \(j=1,2,\dots,p\), which in turn means that Algorithm 1 (main article)
with \(|c_j(\lambda^{(m)})| + \lambda^{(m)}_j - \lambda_j^{(m+1)}\) as
input cannot result in any violations.

From our assumptions we have
\[
  \lvert c_j(\lambda^{(m+1)}) - c_j(\lambda^{(m)})\rvert \leq
  \lvert \lambda_j^{(m+1)} - \lambda_j^{(m)}\rvert.
\]
Using this fact, observe that
\[
  \begin{aligned}
    |c_j(\lambda^{(m+1)})|
     & \leq\lvert c_j(\lambda^{(m+1)}) - c_j(\lambda^{(m)})\rvert + \lvert c_j(\lambda^{(m)})| \\
     & \leq \lambda^{(m)}_j - \lambda^{(m+1)}_j + \lvert c_j(\lambda^{(m)})\rvert.
  \end{aligned}
\]

\hypertarget{proof-proposition3}{%
  \subsection{Proof of Proposition 3}\label{proof-proposition3}}

Let \(c = (\nabla f(\hat\beta(\lambda)))\) and \(\lambda_1 = \lambda_2\) and
assume without loss of generality that \(p=2\) and \(c_1 \geq c_2 \geq 0\).
Recall that the strong rule for lasso discards the \(j\)th predictor whenever
\(c_j < \lambda_1\). There are three cases to consider.

\begin{case}[\(c_2 \leq c_1 < \lambda_1\)]
  \(\mathop{\mathrm{cumsum}}(c - \lambda) \prec 0\), which means both predictors are discarded.
\end{case}
\begin{case}[\(c_1 \geq \lambda_1 > c_2\)]
  The first predictor is retained since \(\mathop{\mathrm{cumsum}}(c - \lambda)_1 > 0\); the
  second is discarded because \(c_2 \leq \lambda\).
\end{case}
\begin{case}[\(c_1 \geq c_2 \geq \lambda_1\)]
  Both predictors are retained since \(\mathop{\mathrm{cumsum}}(c - \lambda) \succeq 0\).
\end{case}
The two results are equivalent for the lasso and thus the strong rule
for SLOPE is a generalization of the strong rule for the lasso.

\hypertarget{algorithms}{%
  \section{Algorithms}\label{algorithms}}

\begin{algorithm}[H]
  \caption{Strong set algorithm}
  \label{alg:strong-alg}
  \begin{algorithmic}
    \State \(\mathcal{V} \gets \varnothing\)
    \State \(\mathcal{E} \gets \mathcal{S}(\lambda^{(m+1)}) \cup \mathcal{T}(\lambda^{(m)})\)
    \Do
    \State compute \(\hat\beta_\mathcal{E}(\lambda^{(m+1)})\)
    \State \(\mathcal{V} \gets\) KKT violations in full set
    \State \(\mathcal{E} \gets \mathcal{E} \cup \mathcal{V}\)
    \doWhile{\(\mathcal{V} \neq \varnothing\)}
    \State \Return \(\hat\beta_\mathcal{E}(\lambda^{(m+1)})\)
  \end{algorithmic}
\end{algorithm}

\begin{algorithm}[H]
  \caption{Previous set algorithm}
  \label{alg:working-alg}
  \begin{algorithmic}
    \State \(\mathcal{V} \gets \varnothing\)
    \State \(\mathcal{E} \gets \mathcal{T}(\lambda^{(m)})\)
    \Do
    \State compute \(\hat\beta_\mathcal{E}(\lambda^{(m+1)})\)
    \State \(\mathcal{V} \gets\) KKT violations in \(\mathcal{S}(\lambda^{(m+1)})\)
    \If {\(\mathcal{V} = \varnothing\)}
    \State \(\mathcal{V} \gets\) KKT violations in full set
    \EndIf
    \State \(\mathcal{E} \gets \mathcal{E} \cup \mathcal{V}\)
    \doWhile{\(\mathcal{V} \neq \varnothing\)}
    \State \Return \(\hat\beta_\mathcal{E}(\lambda^{(m+1)})\)
  \end{algorithmic}
\end{algorithm}

\end{document}